\documentclass[letterpaper, 10 pt, conference]{ieeeconf}  % Comment this line out if you need a4paper

\IEEEoverridecommandlockouts                              % This command is only needed if 
\usepackage{amssymb}
\usepackage{amsmath}
\usepackage{graphicx}
\usepackage{xcolor}
\usepackage{booktabs}
\usepackage{array}
\usepackage{units}
\usepackage{algorithm}
\usepackage{algpseudocode}
\usepackage{wrapfig}
\usepackage{hyperref}
\usepackage{subcaption}
\usepackage{balance}

\title{\LARGE \bf Contrastive Initial State Buffer for Reinforcement Learning}

\author{Nico Messikommer, Yunlong Song, Davide Scaramuzza% <-this % stops a space
\thanks{The authors are with the Robotics and Perception Group, Department of Informatics, University of Zurich, and Department of Neuroinformatics, University of Zurich and ETH Zurich, Switzerland (\protect\url{http://rpg.ifi.uzh.ch}. {\tt\footnotesize nmessi@ifi.uzh.ch}).
This work was supported by the European Union’s Horizon 2020 Research and Innovation Programme under grant agreement No. 871479 (AERIAL-CORE) and the European Research Council (ERC) under grant agreement No. 864042 (AGILEFLIGHT).
}
}

\usepackage[absolute]{textpos}
\begin{document}

% ============= Command to write a header to say "paper accepted at such conference"
\definecolor{somegray}{rgb}{0.5, 0.5, 0.5}
\newcommand{\darkgrayed}[1]{\textcolor{somegray}{#1}}
% The position is defined in absolute coords. Thus, if the page has another width,
% both numbers (4, 0.7) need to be adjusted. The argument {8} specifies the width of the textblock.
\begin{textblock}{11}(2.5, 0.5)
\begin{center}
\darkgrayed{This paper has been accepted for publication at the \\
IEEE International Conference on Robotics and Automation (ICRA), Yokohama 2024. ©IEEE}
\end{center}
\end{textblock}
% ============= 

\maketitle
\thispagestyle{plain}
\pagestyle{plain}

%%%%%%%%%%%%%%%%%%%%%%%%%%%%%%%%%%%%%%%%%%%%%%%%%%%%%%%%%%%%%%%%%%%%%%%%%%%%%%%%
\begin{abstract}
In Reinforcement Learning, the trade-off between exploration and exploitation poses a complex challenge for achieving efficient learning from limited samples. 
While recent works have been effective in leveraging past experiences for policy updates, they often overlook the potential of reusing past experiences for data collection.
Independent of the underlying RL algorithm, we introduce the concept of a \emph{Contrastive Initial State Buffer}, which strategically selects states from past experiences and uses them to initialize the agent in the environment in order to guide it toward more informative states. 
We validate our approach on two complex robotic tasks without relying on any prior information about the environment: (i)  locomotion of a quadruped robot traversing challenging terrains and (ii) a quadcopter drone racing through a track.
The experimental results show that our initial state buffer achieves higher task performance than the nominal baseline while also speeding up training convergence.
\end{abstract}

\vspace{6pt}
\noindent\textbf{Multimedia Material}
A video is available at \url{https://youtu.be/RB7mDq2fhho} and code at \url{https://github.com/uzh-rpg/cl_initial_buffer}

\section{Introduction}
\label{sec:introduction}

In Reinforcement Learning (RL), achieving efficient learning from limited samples has been a central pursuit~\cite{arulkumaran2017deep, chua2018deep}. 
The trade-off between exploration and exploitation lies at the heart of this field, posing a complex problem for researchers and practitioners alike. 
The paradigm of utilizing past experiences to guide future actions has emerged as an effective strategy, leading to the development of techniques such as experience replay~\cite{schaul2015prioritized, lin1992self, mnih2015human}.
One common aspect of these techniques is the reliance on a replay buffer, a structure that stores past experiences. However, these methods primarily focus on reusing past experiences for updating the policy, overlooking a potentially significant factor: reusing past experiences for data collection.
Traditionally, agents in benchmark tasks are often initialized randomly based on a predefined state distribution.
However, many robotic tasks present a considerable challenge for exploration in reinforcement learning since prohibitive amounts of exploration are required to reach certain states and receive some learning signal.
Let's consider a task where a humanoid robot needs to learn how to perform a backflip. 
In this task, directly initializing the robot with initial states drawn from a successful backflip trajectory would significantly reduce the required number of environment interactions compared to starting the robot from the ground. 
This approach leverages prior knowledge about the desired behavior and allows the robot to collect more relevant and informative training data from the beginning. 
As a result, the movements necessary for successful execution are learned much faster compared to initializing the robot from the ground.
However, in most cases, a \emph{successful} trajectory, or even complete information about the environment, is not available beforehand.

\begin{figure}[t!]
\centering
\includegraphics[width=0.45\textwidth]{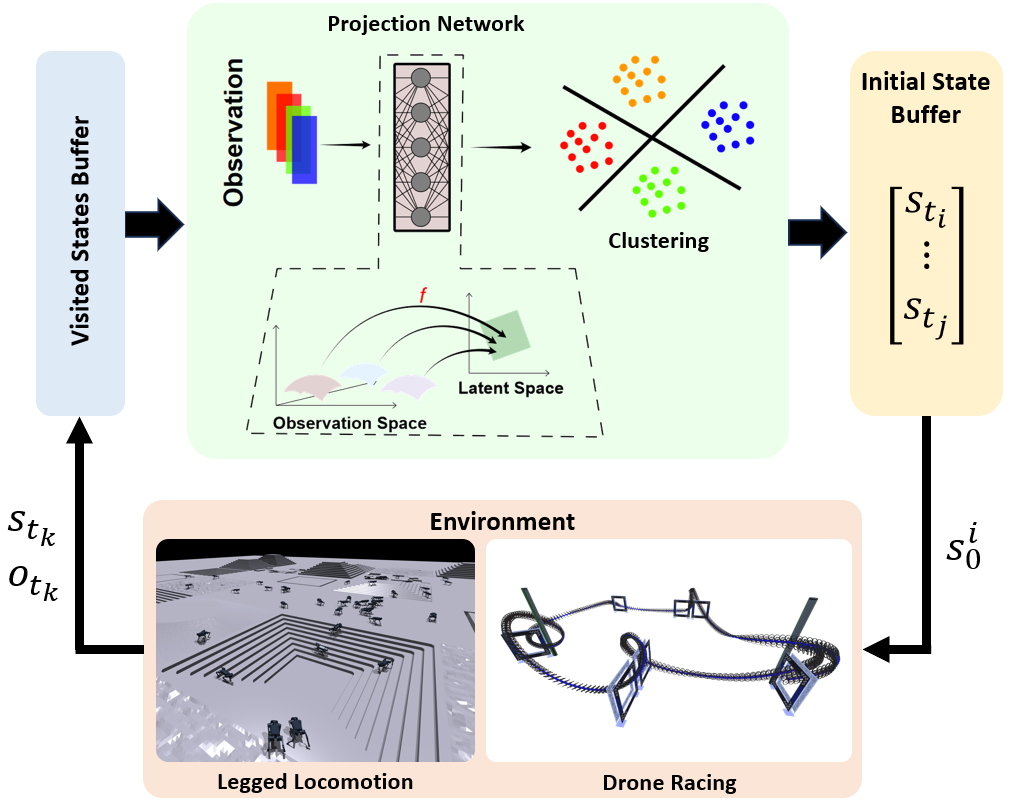}
\caption{
\textbf{Initial State Buffer for Reinforcement Learning.}
Our method uses a network to project observations $o_{t_k}$ to an embedding space, in which we apply K-means clustering.
In a next step, we add states $s_{t_i}$ close to the cluster center to an Initial State Buffer, which sets the initial states of the robot in the environment $s^i_0$ during the roll-out.
}
\label{fig:method_overview}
\vspace{-4pt}
\end{figure}

To achieve the benefits of sampling from an optimal trajectory but without any prior knowledge of the task or environment, we use the concept of \emph{Initial State Buffer} (ISB), which leverages past experiences for data collection.
To optimally select states for the ISB, we propose a novel \emph{Contrastive-Learning Buffer} (CL-Buffer), which maps observations to an embedding space with a learned projection, grouping together states with similar task-relevant experiences, see Fig.~\ref{fig:method_overview}.
The projection is trained in an unsupervised manner, ensuring that states experiencing similar changes in their value function after policy-gradient updates are in close proximity in the embedding space.
By using the distance in the embedding space, a diverse set of states featuring a wide range of experiences can be added to the ISB.
We evaluate the effectiveness of our approach on two complex robotics tasks without relying on any prior information about the environment: (i) a quadruped robot learning to walk from scratch on challenging terrains and (ii) a quadrotor flying through a challenging race track.
Our results show that, for the same number of iterations, the introduction of the CL-Buffer leads to overall faster training convergence along with 18.3\% higher task performance for the quadruped walking task.
Additionally, for the drone racing task, our buffer achieves a success rate of 0.9, in contrast to the baseline without ISB, which has a success rate of 0.2.

\section{Related Work}
\label{sec:related}
Leveraging past experiences for policy improvement has been studied extensively in the literature.
Notably, experience replay~\cite{schaul2015prioritized, lin1992self, mnih2015human, andrychowicz2017hindsight} via a replay buffer plays a key role in the off-policy setting. 
However, standard off-policy RL methods can fail in practice due to the overestimation of values induced by the distributional shift between the collected data and the learned policy.
Moreover, on-policy algorithms suffer from low sample efficiency since they cannot leverage past experiences for policy improvement. 

Researchers are increasingly focusing on enhancing sample efficiency by accumulating more valuable data. 
These approaches exploit domain knowledge, as demonstrated in studies by~\cite{hwangbo2019learning, peng2018deepmimic, song2021autonomous, rudin2021learning}, or leverage reference demonstrations~\cite{salimans2018arxiv, peng2018sfv} to initialize the robot with valid initial states.
For example, in ~\cite{hwangbo2019learning}, the initial state of the robot is sampled either from a previous trajectory or a random distribution, with equal probability. 
Similarly,~\cite{song2021autonomous} proposed using an initial state buffer to store valid states in a drone racing task via a heuristic state machine. 
Compared to our tackled setting, those methods rely on prior knowledge of either the task or the environment, e.g., the race track.
For goal-conditioned tasks, \cite{florensa2017reverse} proposed using reverse curriculum generation, which teaches an agent to reach the goal from starting positions that are progressively distant from the target.
In \cite{ecfoffet21nature}, an archive of visited states is constructed and used to robustify the policy by returning to specified states from the archive. 
Once the agent reaches the selected archive state, it further explores its environment from there.
However, the states are discretized by downscaling images or discretizing robot states, which quickly becomes infeasible for high-dimensional state space common in robotics tasks.
In contrast, we propose a contrastive learning embedding combined with k-means to discretize the visited states adaptively to the capabilities of the agent.
The sampling of states is also used in other fields, e.g., sampling-based motion planning, where supervised learning from optimal trajectories~\cite{ichter2018learning} or Gaussian Mixture Models~\cite{huh2017adaptive} can guide the optimization.
Finally, exploring the environment can also be induced by directly designing RL agents with intrinsic motivation through adapting the reward formulation~\cite{bellemare2016neurips, burda2018iclr, badia2019iclr}. 
Since our proposed method selects initial states for the roll-out without any assumption on the underlying RL agent, our approach can be applied together with these approaches.

\section{Preliminaries}
\label{sec:preliminaries}
General robot tasks solved by Reinforcement Learning (RL) algorithms can be modeled by a \textit{Markov decision process (MDP)} which is described by the tuple $M= (S, A, p, p_0, R, \gamma)$ with state space $S$, action space $A$, transition probabilities between states $p$, initial state distribution $p_0$ and rewards $R$ with discount factor $\gamma$.
While the MDP defines the specific task setting, a stochastic control policy $\pi: S \times A \rightarrow \mathbb{R}$ represents a strategy for the robot to navigate in the given setting by proposing actions $a\sim \pi(s)$ that the robot should take in each state $s \in S$.
Therefore, starting from an initial state distribution $s_0\sim p_0$, a policy induces a probabilistic trajectory of states and actions $\tau = \{s_0, a_0, s_1, a_1, ..., s_T, a_{T-1}\}$. 
As shorthand, we will write that $\tau \sim \pi$ to denote this relationship. 
When starting at a specific fixed state $s_0$, the possible states and actions reached by a policy will be denoted as $\tau \sim \pi | s_0$.
Following the control policy $\pi$, the expected future return achieved by the robot at the current state $s_i$ can be expressed with the value function
\begin{equation}
    V^{\pi}(s_i) = \mathbb{E}_{\tau \sim \pi | s_i} \left[ \sum_{t=i}^{\infty} \gamma^t R(s_t, a_t) \right].
\label{eq:expected_return}
\end{equation}
The goal is to find a policy $\pi^*= \arg\max_{\pi}  \mathbb{E}_{s_0\sim p_0}[V^\pi(s_0)]$, which maximizes the expected cumulative return obtained by starting from initial states $s_0 \sim p_0$.
The distribution of initial states is often crucially important in RL since it determines the overall distribution of roll-outs, and may thus affect learning speed and effectiveness. In what follows, we will have a closer look at this distribution and how to leverage it to increase the sample efficiency.

\section{Initial State Buffer}
\label{sec:initial_state_buffer}
One of the main challenges of RL is low sample efficiency, leading to slow training~\cite{ibarzIJRR21}.
This problem is amplified for large spaces $S$, especially when the transition probabilities impede the fast traversal of the state space by the robot. 
To tackle the sample efficiency problem, we instead propose to sample initial robot states from an \emph{Initial State Buffer} (ISB) which we incrementally build from previously visited states throughout training. 
We leverage this buffer to (\textit{i}) reduce the number of initializations of the agent in situations where it must solve already mastered sub-tasks, since this adds training iterations without improving training, and (\textit{ii}) focus the attention on new or hard situations which require multiple training iterations but are rarely visited. 

This greatly diversifies the robot experience leading to enhanced training performance, an observation which was already made in~\cite{rudin2021learning, hwangbo2019learning, peng2018deepmimic, song2021autonomous, yunlong_science}, albeit with prior information about the environment and task. % with which to select diverse and representative initial states. 
By contrast, our ISB can be used in a general setting without domain-specific knowledge. 
It is thus generally applicable to a diverse set of RL algorithms and problems.  

\subsection{Sampling from the ISB}
The overview of the ISB for the specific subvariant of the CL-Buffer is visualized in Fig.~\ref{fig:method_overview}.
Since storing all of the already visited states during the roll-out phases is intractable, we use a rolling buffer of states called the \textit{visited states buffer} $\mathcal{V}=((s^0, a^0), (s^1,a^1),...,(s^k,a^k))$\footnote{Note here states may come from multiple agents, and thus we use superscripts to avoid confusion with states within one rollout.}, which continuously collects states from the active agents and replaces them in a first-in-first-out fashion. 
After each roll-out phase, $N=256$ states from $\mathcal{V}$ are added to the ISB. 
Every time an episode is terminated, a new episode is started by initializing the robot with a probability of $p=0.8$ at a state from the ISB and a probability of $1-p$ at the original start state $s_0$. 
In what follows, we will discuss three variants of ISB, namely the \textit{Random-Buffer}, \textit{Obs-Buffer}, and \textit{Contrastive Learning Buffer} (CL-Buffer). 

\textbf{Random-Buffer} The Random-Buffer simply selects a set $S$ of  $N$ random states from $\mathcal{V}$
\begin{align}
    S &= \text{RandomSample}_N(\mathcal{V})
\end{align}
While this strategy ensures some diversity, it also tends to oversample states in which the robot spends most of the time during the roll-out episode, e.g. simple transition actions.
These states usually also share similar task-relevant experiences.
It is thus important to modify the sampling algorithm to increase the diversity of states, which is addressed by the next ISB variant.

\textbf{Obs-Buffer} 
To ensure efficient learning, it should be avoided to oversample states from $\mathcal{V}$ that are too similar. 
We thus design an Obs-Buffer, which initiates a K-Means clustering with $K=64$ in the observation space based on the cosine similarity and then samples the $N/K$ states closest to the obtained cluster centers.
\begin{align}
    \{C_k\}_{k=1}^K &= \text{KMeansCluster}(\mathcal{V})\\
    S &= \bigcup_{k=1}^K \text{NearestNeighbor}_{N/K}(C_k),
\end{align}
By doing so, frequently visited states sharing the same observations are all assigned to the same cluster and, thus, are not oversampled. 
While this modification alleviates the problem of oversampling, clustering in the observation space tends to group states together that, actually, lead to completely different task outcomes. 
This is because even small changes in the overall state, especially in unstable configurations, e.g., when a robot is about to crash, can have drastically different expected rewards. 
It is thus beneficial to reenvision how ``closeness" is defined when performing clustering, which is discussed next in the \textit{Contrastive Learning Buffer}.

\subsection{Contrastive Learning Buffer}
\label{sec:embedding_space}
The Contrastive Learning Buffer (CL-Buffer) differs from the Obs-Buffer, in that it trains a neural network $f$, here a three-layer MLP with Tanh activations, to map the observation to an embedding space, before clustering is performed. 
However, how this feature space is constructed is non-trivial since it should encode task knowledge to work effectively. 
Intuitively, we want the network to map states to close-by embeddings if they provide similar experiences to the RL agent since these may require the same "skills", i.e. subroutine of actions, to solve.
We use the term experience here to describe, in general, the distribution of tuples containing the observation, action, and reward information acquired at specific states.
For example, let us consider the skill of a quadruped to climb stairs facing the stairs upward.
In this case, states at the lower and upper part of the stairs provide similar experiences to train the robot to climb stairs.
In contrast, letting a quadruped walk backward on a flat surface will provide different experiences, which are less relevant for climbing stairs.
Thus, the states on the stairs should be clustered together, whereas the state on a flat surface should be further away in the embedding space.
To quantify such a "closeness", we argue that states exhibiting a comparable change in the value function based on a policy gradient update contain shared learning experiences.
We formalize this similarity by analyzing the value function increase at those states after one policy gradient step, which updates a policy $\pi_0$ to $\pi_1$
\begin{align}
    \Delta V^{\pi_1}(s) &\doteq V^{\pi_1}(s) - V^{\pi_0}(s) \nonumber \\
                        &= \mathbb{E}_{(s_0, a_0,...)\sim \pi_0} \left[ \sum_{t=0}^{\infty} \gamma^t A_{\pi_1}(s_t, a_t) \right]  \label{eq:exact_policy_improvement}\\
                        &\approx \sum_{t=0}^{\infty} \gamma^t A_{GAE}^{\pi_1}(s_t, R_t, s_{t+1}, R_{t+1}, ...). \label{eq:approx_policy_improvement}
\end{align}
Here we approximate the change of the value function following~\cite{schulman15icml} by using the \textit{advantage function} $A_{\pi_1}(s_t, a_t)$. 

\begin{figure}[ht!]
% \begin{wrapfigure}{r}{0.5\textwidth}
\centering
\includegraphics[width=0.35\textwidth]{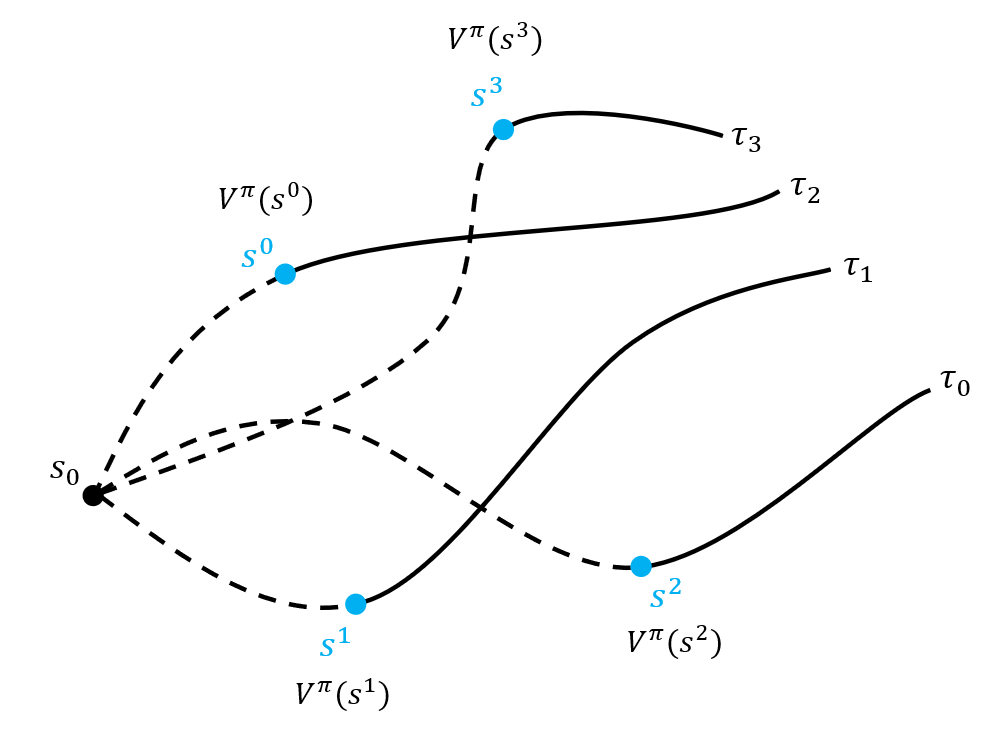}
\caption{
\textbf{sub-MDP.} A standard MDP problem can be divided into multiple sub-MDPs.
}
\label{fig:sub_mdp}
\end{figure}

\textbf{Training the Embedding Network}
At the beginning of each policy update phase, we select a fixed set of states $s^0, s^1, ..s^i$ from the visited state buffer $\mathcal{V}$, see Fig.~\ref{fig:sub_mdp}.
For each of those states, we can define a sub-MDP, which shares all of the same properties as the original MDP, i.e., $M= (S, A, p, R, \gamma)$ except for a different start distribution $p_0$.
At those selected states, we can compute the value function increase $\Delta V^{\pi_1}(s)$ after each policy update.
Since we only have access to the roll-out trajectories obtained with $\pi_0$, we leverage the fact that Eq.~\ref{eq:exact_policy_improvement} uses the expectation over the trajectories $\tau \sim \pi_0$, which we approximate by considering only one sample $\tau$ obtained during the roll-out phase.
Additionally, we approximate the advantage function in Eq.~\ref{eq:exact_policy_improvement} with the \textit{General Advantage Estimator} introduced in \cite{schulman16iclr}.
Eq.~\ref{eq:approx_policy_improvement} enables us to quantify for each policy gradient step the improvement over the original policy, for which we have collected the roll-out samples.
% \end{wrapfigure}
% 
This difference in the value function at different states can then be used to cluster states with similar task-relevant experiences together.
For each policy update step, the predefined states can be ranked according to the approximated value function increase $\Delta V^{\pi_1}(s)$.
The top-k states with the highest increase in the expected return are then used as a positive set $P$, whereas the lowest-k states are used as a negative set $N$.
Additionally, we select the embedding $\textbf{x}_i = f(s_i)$ of a state from the positive sets $P$ as an anchor point.
Similarly to ~\cite{xing2024icra}, we use the anchor point, the positive and negative set to formulate the soft-nearest neighbor loss~\cite{frosst19icml}, which is a version of the InfoNCE loss~\cite{oord2018arxiv}, as follows
\begin{align}
    L_{\text{contra}} &= - \log \frac{\sum_{j \in P} \exp{(-g(\mathbf{x_i}, \mathbf{x}_j) / \tau)}} {\sum_{k \in N \bigcup P} \exp{(-g(\mathbf{x_i}, \mathbf{x}_k) / \tau)}}
\label{eq:info_nce}
\end{align}
The function $g$ computes, in our case, the cosine similarity between two different embedding vectors $x_i, x_j$ while $\tau$ is a temperature parameter.
By backpropagating through the loss, we can update the parameters of the projection network $f$. 
Using this network, we map the states in the visited state buffer $\mathcal{V}$ to the embedding space and then perform K-Means clustering. The sampling procedure thus becomes 
\begin{align}
    \mathcal{X} &= \{x_1,...,x_N\} = f(\mathcal{V})\\
    \{C_k\}_{k=1}^K &= \text{KMeansCluster}(\mathcal{X})\\
    S &= \bigcup_{k=1}^K \text{NearestNeighbor}_{N/K}(C_k),
\end{align}
where we compute the nearest neighbors in the embedding space. %, and sample the states corresponding to the embeddings that are sampled. 
Note that while states are regularly mapped to the embedding space to perform sampling, the network is continually trained throughout RL training, which means that it can learn to shift the focus from states as they no longer change in terms of the value function. This type of clustering thus adds a dynamic behavior to the ISB. 

\newcolumntype{C}[1]{>{\centering}m{#1}}

\section{Experiments}
\textbf{Setup}
We evaluate the different ISB strategies on two robotic platforms: a quadruped and a quadrotor. 
The first task involves command following for a quadruped platform, based on the implementation of~\cite{rudin2021learning}.
The input to the RL agent is the heading direction, which is translated to an angular rotation, and linear velocities commands for the $x$- and $y$-axis expressed in its body frame.
Based on the commands, the quadruped needs to traverse across different terrain, including upwards- and downwards stairs, rough terrain, smooth terrain, discrete terrain with different elevation planes, and flat surfaces.
These terrains are created as square areas of 8m$\times$8m in a simulated IsaacGym~\cite{makoviychuk2021isaac} environment.
During training, the robot is initialized in the middle of the whole environment and needs to explore the other terrain types.
However, during validation, we initialize the robot in the center of all of the different terrain types, which enables us to directly asses how well the robot learned to walk on the different terrains.
These validation runs are conducted throughout the training process in time intervals of 100 roll-out phases.
We perform 2000 training iteration steps with 1024 environments, which delivers the best trade-off between result expressibility and run-time cost.
In the second task, we use the different ISB strategies to train an agent for drone racing, which is a challenging task that requires pushing the drone to its physical limit~\cite{yunlong_science, kaufmann2023champion}.
For this task, we train our agent in the Flightmare simulator~\cite{yunlong2020flightmare}. 
The agent receives as input the relative gate corner of the next gate and the angular and linear velocity of the quadrotor and outputs the collective thrust and the body rates.
Like the quadruped task, we evaluate each method after 100 roll-out phases.
As in~\cite{yunlong_science}, we use 2000 training iterations with 100 environments.

\textbf{Implementation}
We evaluate the different ISB variants: Random-Buffer, Obs-Buffer, and CL-Buffer, as well as the nominal \emph{Vanilla} baseline without an ISB.
To avoid introducing bias to the evaluation of the ISBs, we adopt the same PPO algorithm and reward function design used in~\cite{rudin2021learning} for the quadruped platform and the framework from~\cite{yunlong_science} for the quadrotor. % with only minor changes.
Furthermore, the same environments and training frameworks were shown multiple times to transfer to the real world for the quadruped~\cite{rudin2021learning, frey22iros, miki22science} and for the quadrotor~\cite{yunlong_science, romero2024icra}.
This confirms the validity of the simulation for real-world robotic applications.
We design a simple pre-selection of states to be added to the visited state buffer $\mathcal{V}$.
For the quadrupedal locomotion, we exclude states less than 15 timesteps away from their episode start and states with an accumulated reward below zero since this indicates that the command following is not performed.
The ablation experiments in Fig.~\ref{fig:train_perform} (d) conducted with our CL-Buffer show that combining both filtering steps leads to the best training performance throughout the training.
For drone racing, we only add states to the buffer, which are visited in trajectories initialized at the original start state to avoid diverging from the race track.
Generally, our method is not significantly impacted by parameters like cluster numbers and buffer size for visited states.
\section{Results}
\label{sec:result}
\subsection{Quadrupedal Locomotion}
% 
% \begin{figure*}[ht!]
% \centering
% \includegraphics[width=0.8\textwidth]{Floats/Figures/Images/training_performance.png}
% \caption{
% \textbf{Quadrupedal Locomotion Comparison.}
% The mean validation performance at different iteration steps obtained with five different training seeds for all of the tested methods.
% % 
% }
% \label{fig:train_perform}
% \end{figure*} 

\begin{figure*}
     \centering
     \begin{subfigure}[b]{0.24\textwidth}
         \centering
         \includegraphics[trim={0 0 0 0.5cm},clip,width=\textwidth]{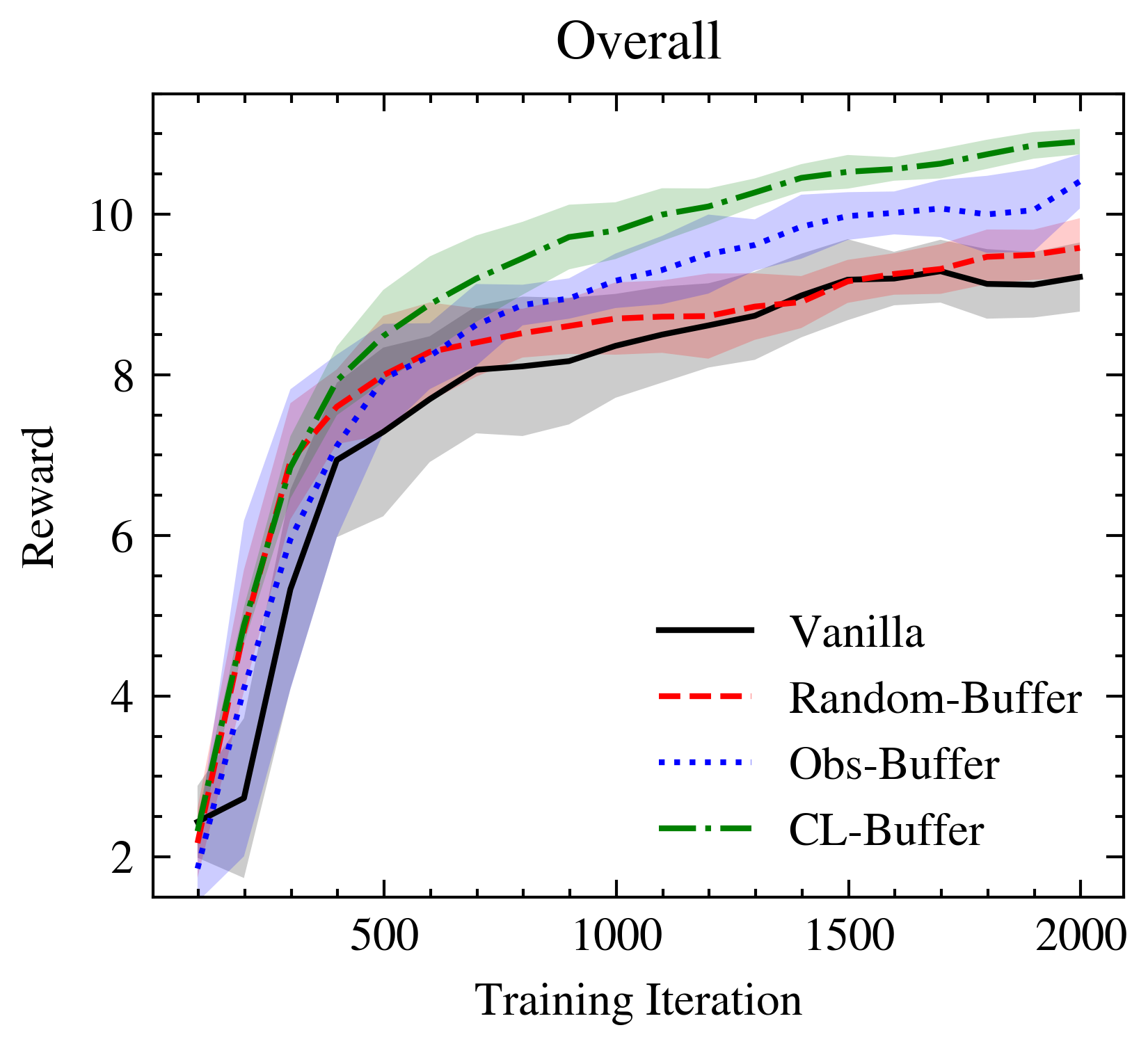}
         \caption{Training Performance}
     \end{subfigure}
     \hfill
     \begin{subfigure}[b]{0.24\textwidth}
         \centering
         \includegraphics[trim={0 0 0 0.5cm},clip,width=\textwidth]{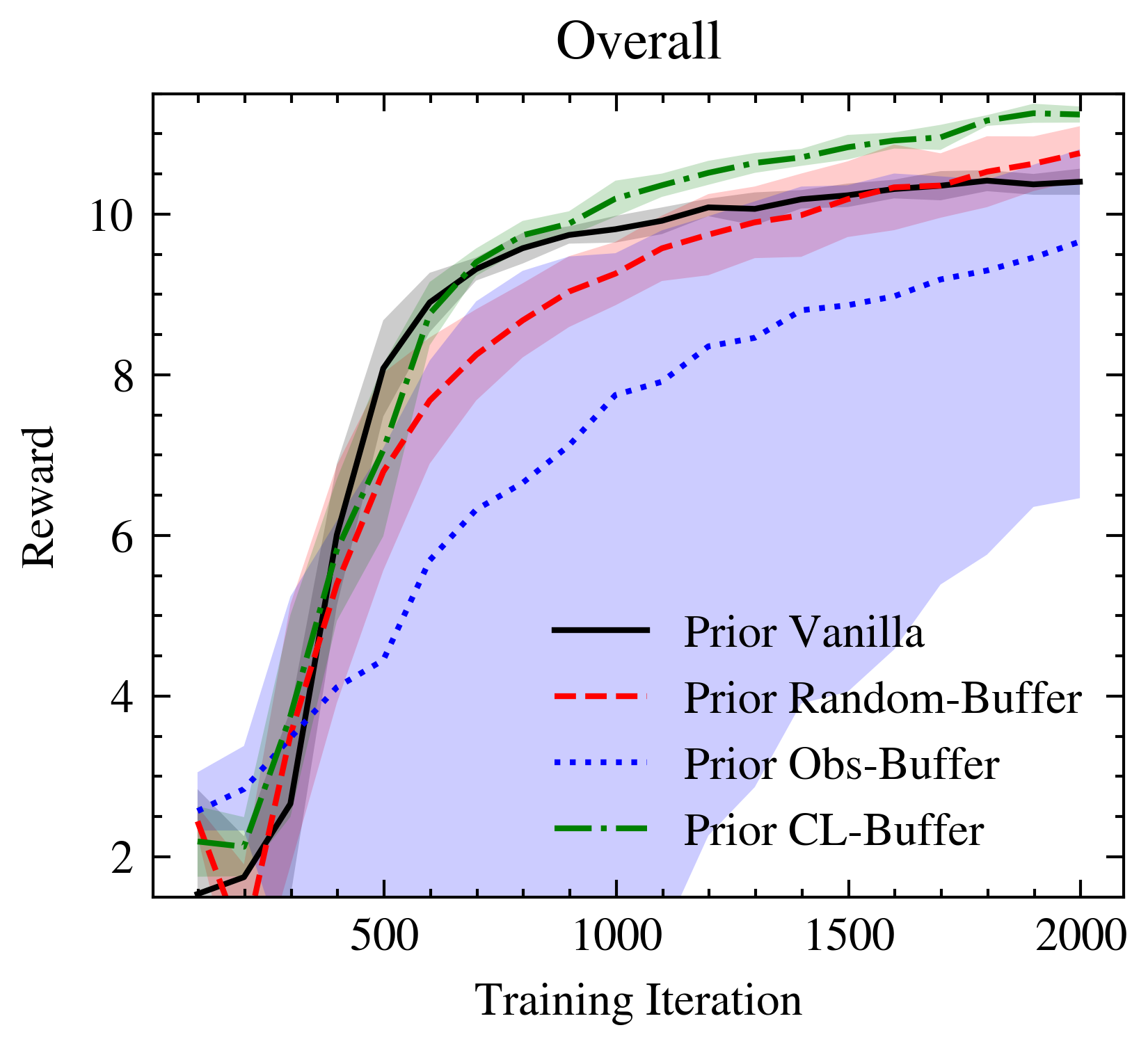}
         \caption{Prior Information}
     \end{subfigure}
     \hfill
     \begin{subfigure}[b]{0.24\textwidth}
         \centering
         \includegraphics[trim={0 0 0 0.5cm},clip,width=\textwidth]{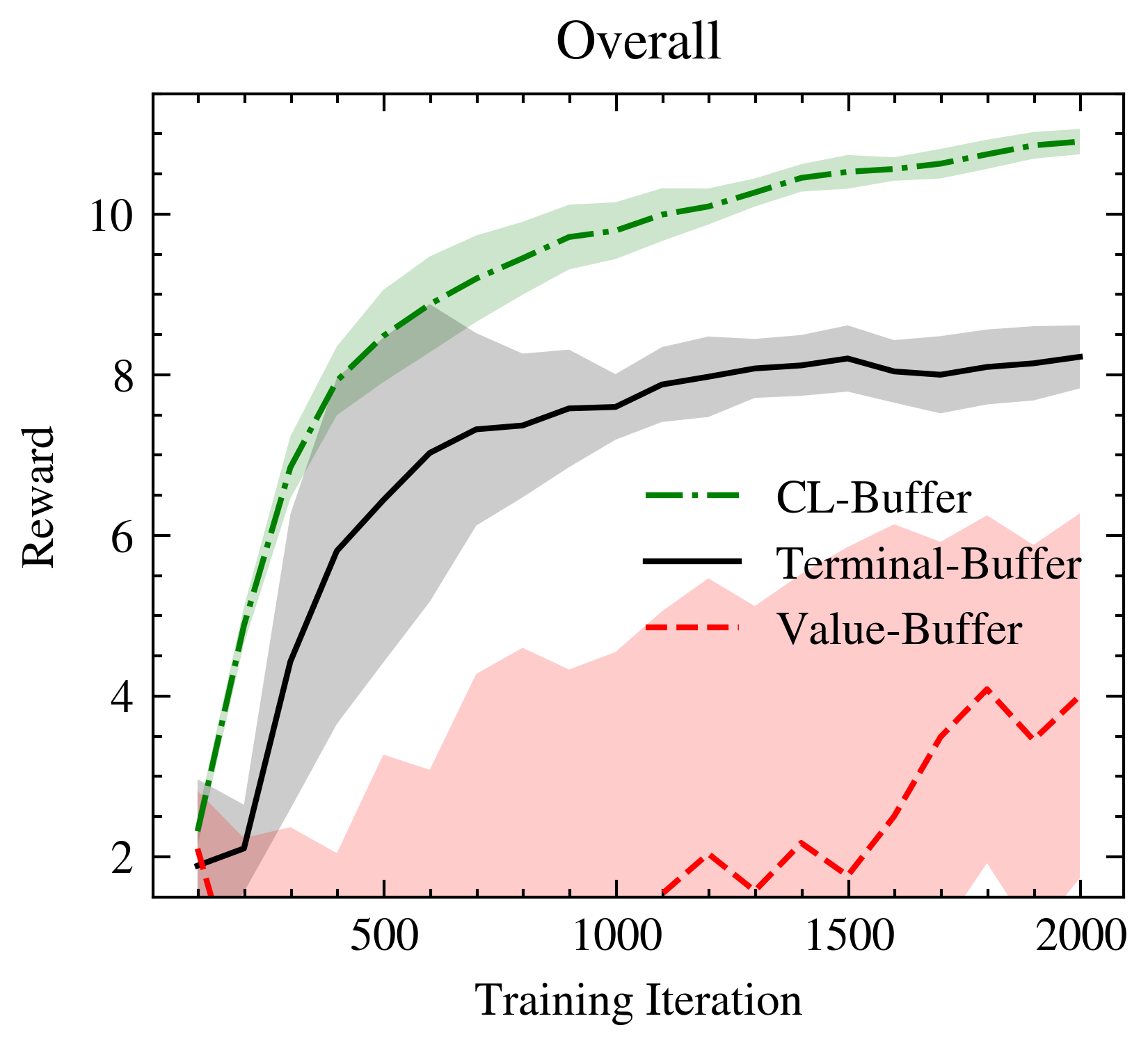}
         \caption{Other Baselines}
     \end{subfigure}
     \hfill
     \begin{subfigure}[b]{0.24\textwidth}
         \centering
         \includegraphics[trim={0 0 0 0.5cm},clip,width=\textwidth]{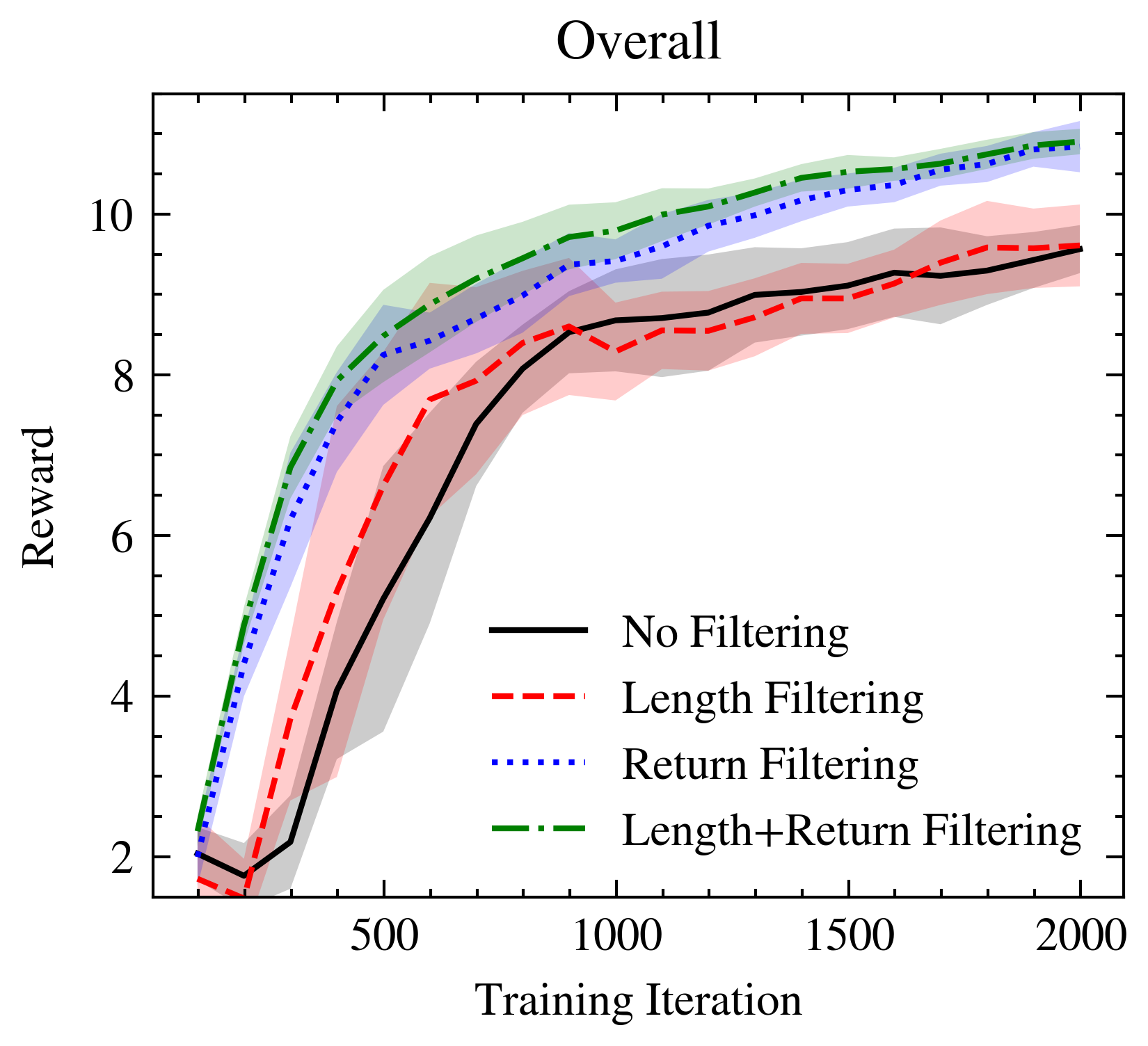}
         \caption{Visited States Filtering}
     \end{subfigure}
\caption{
\textbf{Quadrupedal Locomotion.}
The mean validation performance at different iteration steps obtained with five different training seeds for all of the tested methods.}
\label{fig:train_perform}
\end{figure*}

\begin{table}[ht!]
% \footnotesize
\centering
\begin{tabular}{m{1.8cm}C{1.6cm}C{1.6cm}>{\centering\arraybackslash}m{1.6cm}}
     & \multicolumn{3}{c}{Reward $\uparrow$ } \\
\cmidrule(lr){2-4}
Method  & Overall & Easy & Hard \\
\hline
Vanilla         & 9.21$\pm$0.43           & 10.19$\pm$0.46          & 7.62$\pm$0.44 \\
Random-Buffer   & 9.58$\pm$0.36           & 10.45$\pm$0.35          & 8.14$\pm$0.43 \\
Obs-Buffer      & 10.40$\pm$0.33          & 11.13$\pm$0.30          & 9.19$\pm$0.38 \\
CL-Buffer       & \textbf{10.90$\pm$0.15} & \textbf{11.81$\pm$0.08} & \textbf{9.46$\pm$0.36}
% Old
% Vanilla         & 8.39 & 9.45  & 6.67 \\
% Random          & 9.14 & 10.04 & 7.67 \\
% Observations    & 9.82 & 10.64 & 8.48 \\
% Network         & 9.99 & 10.87 & 8.58
% \hline
\end{tabular}
\caption{
\textbf{Quadrupedal Locomotion.}
The validation performance averaged over five different training seeds for different terrain difficulties according to the parametrization of~\cite{rudin2021learning}.
}
\label{tab:final_training_perform}
\end{table}
\textbf{Performance of CL-Buffer}
Fig.~\ref{fig:train_perform} (a) shows the validation performance of the tested methods at the different training iteration steps.
Since we initialize the robot at multiple terrains during validation, we can directly assess the learned capabilities for different terrain difficulties.
It can be observed that initializing the states using an ISB speeds up the learning at the start of the learning, as confirmed by the higher validation performance achieved by all of the ISB methods until the training iteration 1000.
However, in the later stages of the training, the Random-Buffer converges to the same performance as the Vanilla approach. 
% % 
This confirms the importance of strategically adding states to the ISB.
% % 

\begin{figure*}[ht!]
\centering
\includegraphics[width=0.99\textwidth]{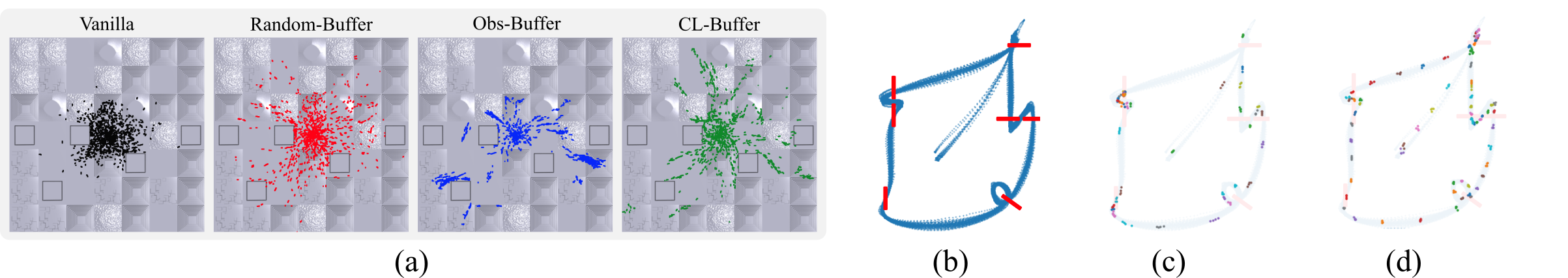}
\caption{
\textbf{State Distribution.}
(a) The quadruped states during a roll-out phase in the middle of the training in a top-down view.
(b) For flying through a race track, if the agent can not fly yet through the complete racetrack, (c) the initial state clusters from the CL-Buffer are located around the struggling gate.
(d) Once the agent can finish the racetrack, the CL-Buffer clusters are more spread out while still focusing on the difficult parts around the gates and less on the straight lines.
% 
% The different colors represent the different tested methods.
% 
% The \emph{Vanilla} approach has a limited exploration radius since it always initializes the roll-out trajectories in the middle of the environment.
% % 
% Whereas, the \emph{Random} sampling increases the state distribution in an even way, which leads to the sampling of irrelevant states.
% % 
% The \emph{Obs-Buffer} creates clustered states, which can be explained by the difficulty of selecting states based on a distance in the observations space.
% % 
% In contrast, our \emph{CL-Buffer} leads to a diverse state distribution with a large exploration radius using the distance in a learned embedding space.
% 
}
\label{fig:topdown}
\end{figure*}

% 
In terms of final validation performance, our CL-Buffer outperforms Obs-Buffer, the closest ISB method, by 4.8\% and improves the performance compared to the Vanilla baseline by 18.3\%, as also reported in Tab.~\ref{tab:final_training_perform}.
This confirms the effectiveness of our proposed state selection based on a learned projection network.
It needs to be emphasized that the performance increase is achieved without changing the underlying RL algorithm, which shows the potential of our proposed ISB for general RL tasks.
In Fig.~\ref{fig:topdown} (a), we visualize the state distributions for the tested methods during a roll-out phase performed in the middle of the training.
The quadruped states are visualized in a top-down view of the environment using different colors for different methods.
The figure shows that the Vanilla approach has a limited exploration radius since it always initializes the trajectories in the middle of the environment.
As can be further observed, the Random-Buffer increases the state distribution evenly, which leads to the sampling of redundant experiences.
To select a more diverse set of initial states with more relevant experiences, the Obs-Buffer uses observations as distance metrics in the K-Means clustering.
However, the observation space can cluster uncorrelated dimensions, making it an unsuitable metric for selecting diverse states.
This can lead to states being selected for the ISB, which share similar experiences and are close in the environment, as visualized in Fig.~\ref{fig:topdown} (a).
Instead, our CL-Buffer uses an embedding space that clusters states with similar experiences.
The result is a diverse state distribution, leading to an improved performance confirmed by the results.
Furthermore, we evaluate the methods if prior information about the environment is available by initializing the agents at the center of all terrains during the training.
As can be observed in Fig.~\ref{fig:train_perform} (b), the Prior CL-Buffer leads to the best training performance and, compared to the Prior Vanilla, also has a steeper reward curve at the end of the training, indicating that the performance gain could even further increase.
In comparison, the Prior Random-Buffer struggles in selecting useful states if the training data already contains a widespread of situations present in the tested environment.
For the Prior Obs-Buffer, the uniform terrain initialization can result in initial state clusters of excessively high difficulty, causing some training runs to collapse, which explains the high variance of the reward curve.
If we compare the Prior Vanilla with our CL-Buffer in Fig.~\ref{fig:train_perform} (a), which does not have the significant advantage of having access to environment information, the Prior Vanilla only performs similarly and even slightly worse than our proposed CL-Buffer.
Furthermore, we evaluate two other simple ISB strategies in Fig.~\ref{fig:train_perform} (c): a Terminal-Buffer, which selects states close to the terminal states of previously conducted roll-outs as initial states, and a Value-Buffer, which selects states from the visited state buffer with the highest value predicted by the estimated value function. 
Both strategies lead to an underperforming training performance.

\textbf{Analysis of CL-Buffer Clusters}
\begin{figure}[ht!]
\centering
\includegraphics[width=0.43\textwidth]{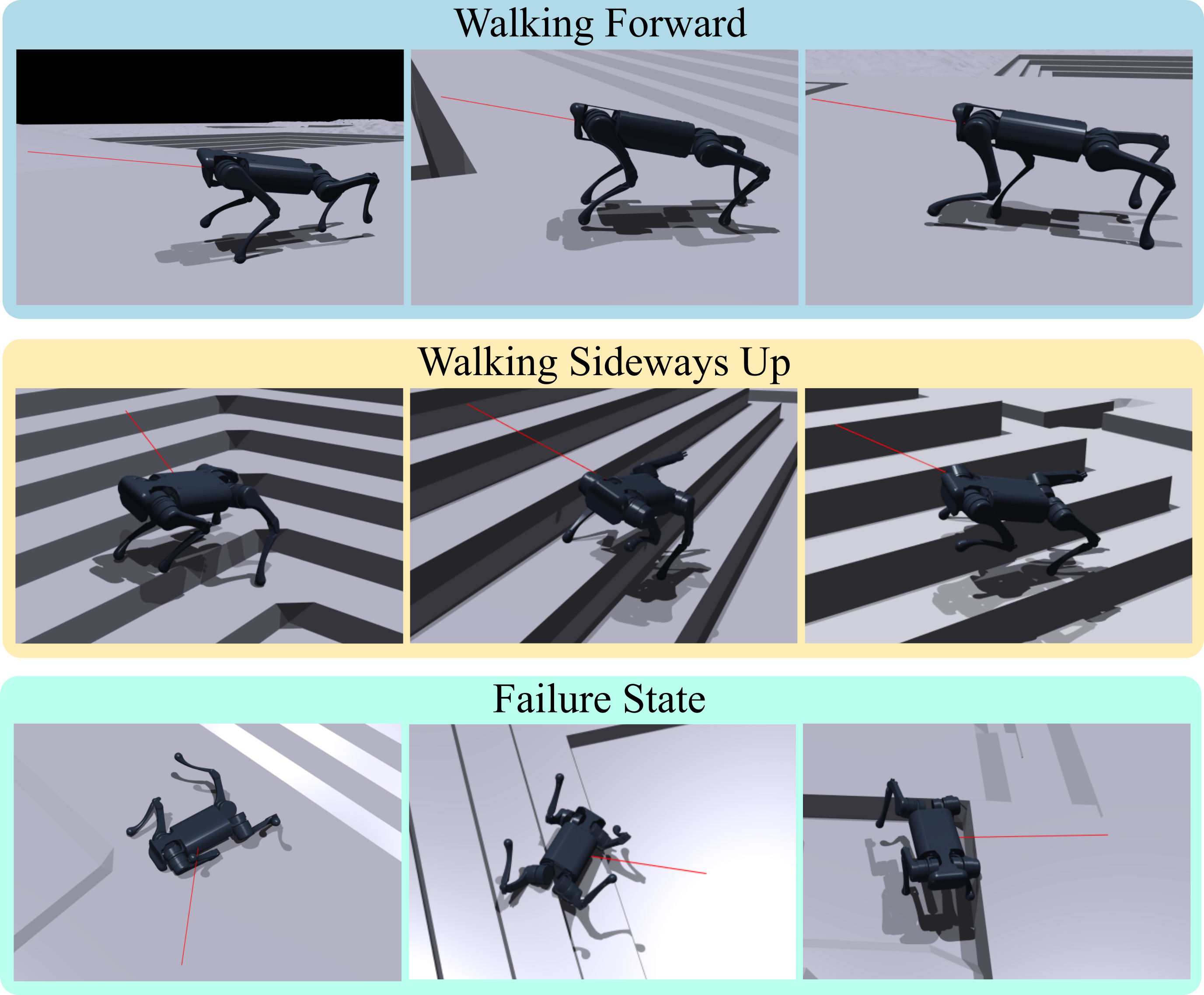}
\caption{
\textbf{Cluster State Visualization.}
Our proposed projection network can be used to cluster the embeddings of different states according to the corresponding experience.
It can be observed that each cluster represents a specific skill, i.e., \emph{Walking Forward}, \emph{Walking Sideways Up}, and \emph{Failure State}.
}
\label{fig:state}
\end{figure} 
To provide more insights into the state clustering of the learned embedding space, three samples closest to three cluster centers are visualized in Fig.~\ref{fig:state}, which shows the state of the quadruped and the given command visualized as a red line.
The visualized clusters can be characterized into different skills, e.g., "Walking Forward," "Walking Sideways Up," and "Failure State."
Notably, these skill-based clusters evolved during training and were not directly enforced.
This shows that our CL-Buffer selects a diverse set of states featuring different experiences for different skills and thus increases the training convergence and final performance, as confirmed by the validation performance.
Furthermore, the capability of our projection network to cluster states in an unsupervised fashion into skills relevant to the given task and adaptively to the current policy throughout the training opens up the door for various applications of the proposed clustering.
For example, the projection network can be used to design training environments in a curriculum-learning fashion by interpolating between visited states, similarly done as in previous works in Deep Learning~\cite{zhang2018iclr} and RL~\cite{sander2022ldcc, lin2021icra}.
We believe that our projection technique will provide the tool for many future applications.

\subsection{Drone Racing}
\begin{figure}[ht!]
\centering
\includegraphics[width=0.48\textwidth]{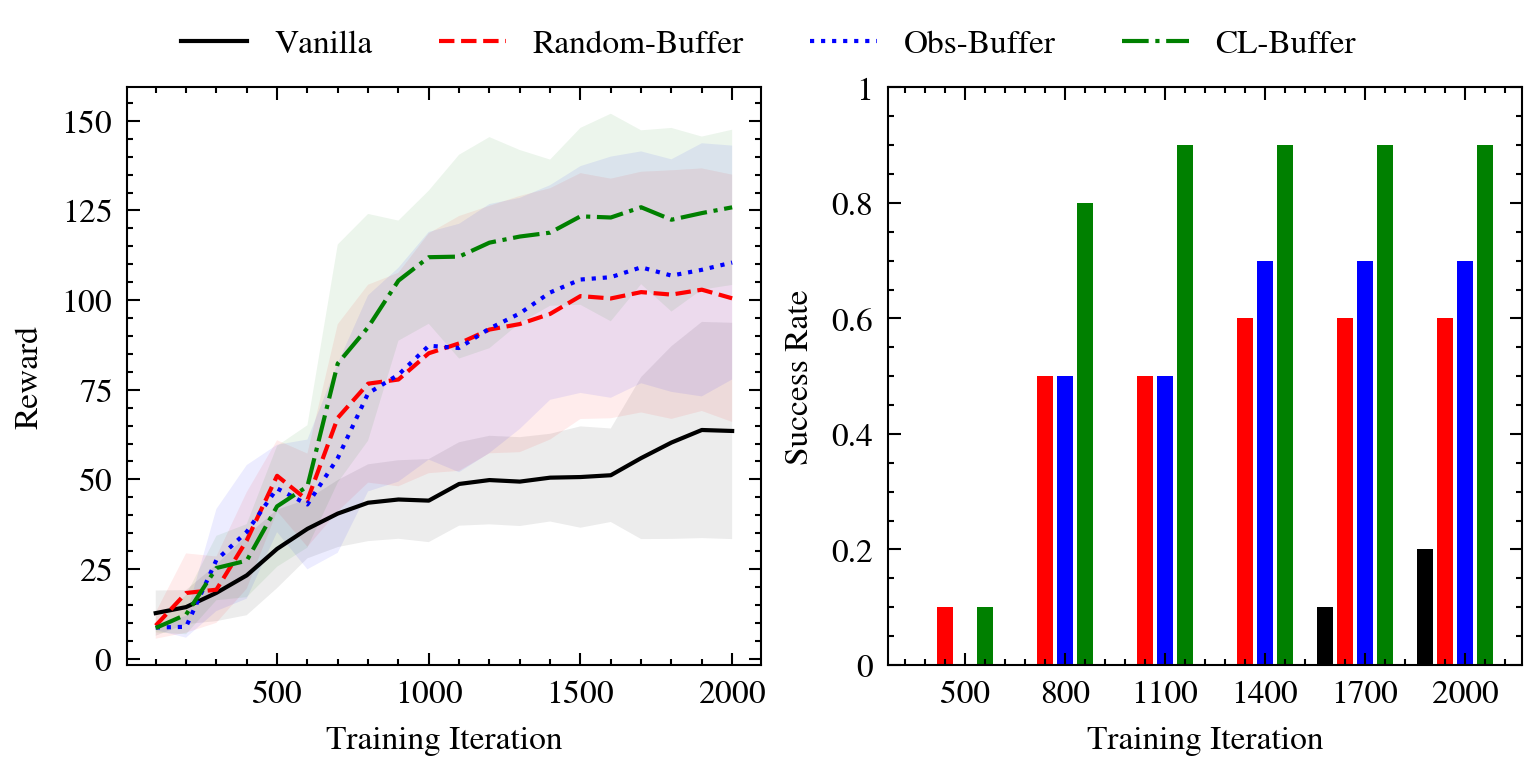}
\caption{
\textbf{Drone Racing.}
The mean validation reward (left), and the success rate (right) obtained with ten different seeds.
}
\label{fig:racing_results}
\end{figure}

\textbf{Performance of CL-Buffer}
To show that the ISB framework works on other environments and platforms without any major changes, we evaluate the methods for drone racing.
In drone racing, the objective is to train a policy to pass through a sequence of gates in minimum time, as visualized in a top-down view in Fig.~\ref{fig:topdown} (b). 
Using RL to address this problem has one inherent challenge: delayed gratification. 
Progress in early gates is critical to reaching and exploring later gates. The agent may need to pass through gates 1 and 2 before seeing gate 3. 
Thus, a wrong action in gate 1 might not manifest its negative consequence until gate~3. This long causal chain makes the credit assignment difficult.
As shown in Fig.~\ref{fig:racing_results}, the experimental results clearly show the benefit of using an ISB.
Our proposed CL-Buffer leads to the fastest training convergence and the highest success rate, indicating how many of the agents trained with different seeds can complete one round of the race track.
The achieved lap times of successful passes are comparable for the different ISB while the CL-Buffer leads to more successful passes.
\textbf{Analysis of CL-Buffer Clusters}
The states found by the CL-Buffer are visualized in Fig.~\ref{fig:topdown} (c) and (d) as dots with different colors, while initial states corresponding to the same cluster center have the same color and are close in proximity.
The top-down view in Fig.~\ref{fig:topdown} (c) shows that if the agent has not mastered flying through the entire racetrack, most of the cluster samples are located on the struggling gate, which is, in this case, the last gate.
Once the agent can finish the racetrack, the cluster centers are more spread out while focusing on the difficult parts around the gates, see Fig.~\ref{fig:topdown} (d).
This shows the evolution ability of our proposed CL-Buffer to adapt to the capabilities of the current trained agent.
% 

% \begin{figure*}[ht!]
% \centering
% \includegraphics[width=0.99\textwidth]{Floats/Figures/Images/racing.png}
% \caption{
% \textbf{Drone Racing Comparison.}
% (Left): The mean validation reward evaluated at different training iterations. The curves are obtained with ten different training seeds.
% (Middle): The lap time evaluated at different training iterations.
% (Right): Success rates. 
% Overall, the CL-buffer can lead to faster convergences as well as higher success rates. 
% % The left plot shows the mean validation performance at different iteration steps obtained with ten different training seeds for all of the tested methods.
% % The lap times (middle) and the success rate of completed tracks (right) achieved by the agents at different training iterations show that the CL-Buffer leads to equivalent racetimes compared to the other buffer methods. 
% % However, the CL-Buffer leads a higher number of agents able to complete the race track early on in the training.
% }
% \label{fig:racing_results}
% \end{figure*} 

\section{Conclusion} 
\label{sec:conclusion}
This work evaluated the concept of Initial State Buffers (ISB), which strategically select states from past experiences to initialize the robot in the environment in order to guide it toward more informative states.
To ensure that the ISB contains states with diverse task-relevant experiences, we proposed a novel Contrastive Learning Buffer (CL-Buffer), which relies on a learned projection to an embedding space in which close-by states share similar experiences.
Our results show that the CL-Buffer leads to overall faster training convergence along with 18.3\% higher task performance for the quadruped walking task.
Additionally, for the drone racing task, our buffer achieves a success rate of 0.9, in contrast to the baseline without ISB, which has a success rate of 0.2.
By leveraging task and environment information, future work can extend our CL-Buffer to consider a prioritized sampling from the initial state buffer.
% 

%%%%%%%%%%%%%%%%%%%%%%%%%%%%%%%%%%%%%%%%%%%%%%%%%%%%%%%%%%%%%%%%%%%%%%%%%%%%%%%%

\balance 

\bibliographystyle{IEEEtran}
{
\bibliography{references}
}

\end{document}